\documentclass[journal]{IEEEtran}
\usepackage{makecell}
\usepackage{multirow}
\usepackage[colorlinks]{hyperref}

\usepackage{epsfig}
\usepackage{graphicx}
\usepackage{amsmath,amssymb} 
\usepackage{color}
\usepackage{array}
\usepackage{bbm}

\usepackage{algorithm}
\usepackage{algorithmicx}
\usepackage{algpseudocode}
\usepackage{enumitem}
\usepackage{graphicx}
\usepackage{amssymb}
\usepackage{pifont}
\usepackage{amsfonts}
\usepackage{booktabs}
\usepackage{siunitx}
\usepackage{subfigure}
\usepackage{graphicx} 
\usepackage{epsfig}
\usepackage{float}
\usepackage{lipsum}

\makeatletter
\def\eg{\emph{e.g.},~} 
\def\ie{\emph{i.e.},~} 
\def\etc{\emph{etc.}} 
\makeatother

\begin{document}
%

\title{Boosting Semi-Supervised Face Recognition with Noise Robustness}
%
%
%

\author{Yuchi Liu, Hailin Shi,~\IEEEmembership{Member,~IEEE}, Hang Du, Rui Zhu, Jun Wang,\\ Liang Zheng,~\IEEEmembership{Senior Member,~IEEE} and
        Tao Mei,~\IEEEmembership{Fellow,~IEEE}
\thanks{Y. Liu and L. Zheng are with the School of Computer Science and Engineering, Australian National University, Australia. E-mail: yuchi.liu@anu.edu.au, liang.zheng@anu.edu.au.

H. Shi, H. Du, J. Wang and T. Mei are with the JD AI Research, JD Group, Beijing, China. E-mail: shihailin@jd.com, duhang@shu.edu.cn, wangjun49@jd.com, tmei@live.com.

R. Zhu is with the Department of Computer Science and Engineering, Chinese University of Hong Kong, Shenzhen, China. E-mail: ruizhu@link.cuhk.edu.cn


}
}

%

\maketitle

\begin{abstract}
Although deep face recognition benefits significantly from large-scale training data, a current bottleneck is the labelling cost. A feasible solution to this problem is semi-supervised learning, exploiting a small portion of labelled data and large amounts of unlabelled data. The major challenge, however, is the accumulated label errors through auto-labelling, compromising the training.
In this paper, we present an effective solution to semi-supervised face recognition that is robust to the label noise aroused by the auto-labelling.
Specifically, we introduce a multi-agent method, named GroupNet (GN), to endow our solution with the ability to identify the wrongly-labelled samples and preserve the clean samples.
We show that GN alone achieves the leading accuracy in traditional supervised face recognition even when the noisy labels take over 50\% of the training data.
Further, we develop a semi-supervised face recognition solution, named Noise Robust Learning-Labelling (NRoLL), which is based on the robust training ability empowered by GN.
It starts with a small amount of labelled data and consequently conducts high-confidence labelling on a large amount of unlabelled data to boost further training.
The more data is labelled by NRoLL, the higher confidence is with the label in the dataset.
To evaluate the competitiveness of our method, we run NRoLL with a rough condition that only one-fifth of the labelled MSCeleb is available and the rest is used as unlabelled data.
On a wide range of benchmarks, our method compares favorably against the state-of-the-art methods. ~\footnote{Our code will be available at \href{https://github.com/liuyvchi/NROLL}{https://github.com/liuyvchi/NROLL}.}
\end{abstract}


%
\IEEEpeerreviewmaketitle

\section{Introduction}
\label{sec_intro}
\begin{figure}[t] 
    \centering
    \begin{center}
        
        \includegraphics[width=0.95\linewidth]{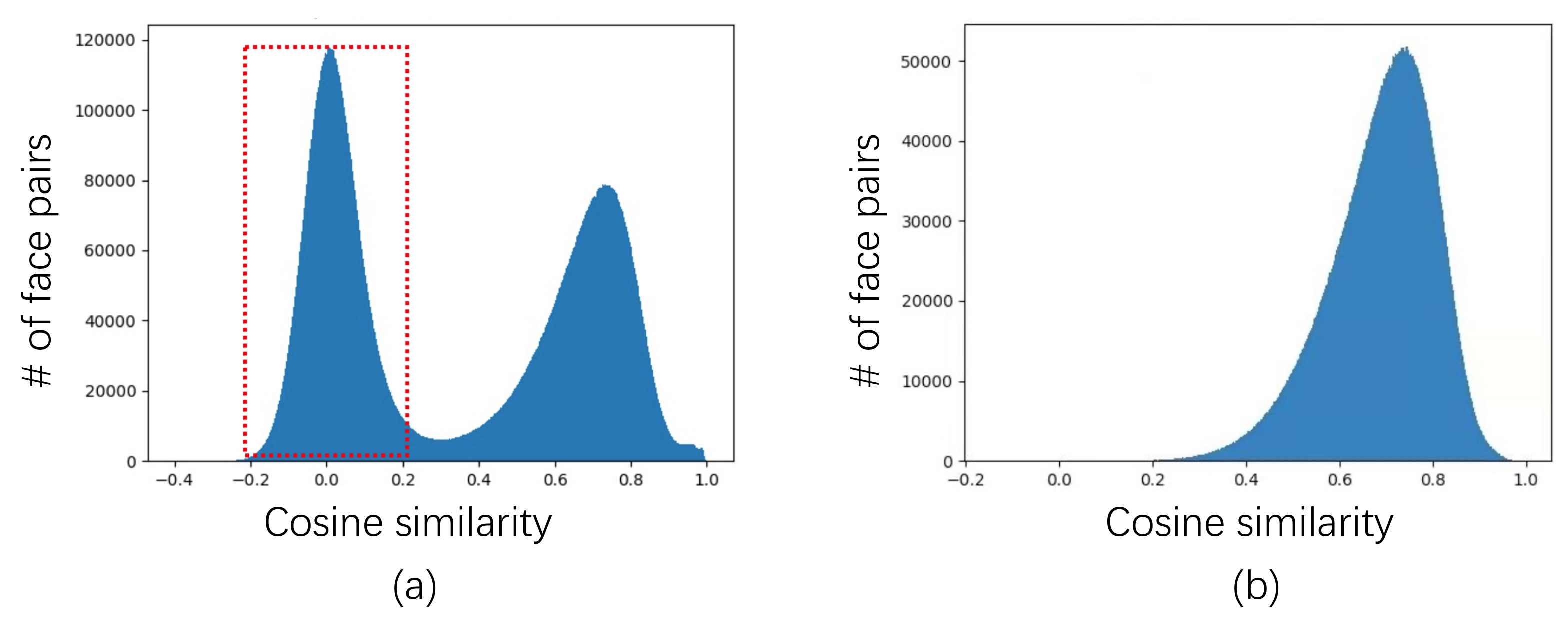}
      
        \caption{Similarity histogram of intra-class face pairs in MSCeleb~\cite{guo2016ms} (a) and its manually-cleaned version by TrillionPairs~\cite{deepglint} (b). The two peaks in (a) indicate the large portion of noisy labels, whereas the manual correction removes the noisy-label samples and leaves one peak in (b).} 
        \label{fig: MSC-R-intra} 
    \end{center}
    \vspace{-1em}
\end{figure}

\IEEEPARstart{T}{he} key points to high-performance deep face recognition include large-scale training data, deep convolutional neural networks (CNN), and advanced training methods. In recent years, many great works have studied in training with evolving objectives, such as SphereFace~\cite{liu2017sphereface}, ArcFace~\cite{deng2019arcface}, CosFace~\cite{wang2018cosface}, \etc These methods are well developed in the manner of supervised learning, which need large amount of labelled training data to bring their advantage. Therefore, many large-scale datasets have been proposed for deep face recognition as well, including CASIA-Webface~\cite{Dong2014Learning}, MegaFace~\cite{kemelmacher2016megaface}, MSCeleb~\cite{guo2016ms}, VGGFace~\cite{parkhi2015deep}, VGGFace2~\cite{cao2018vggface2}. Whereas the data scale keeps growing with increasing number of identities, the workload and complexity of data labelling also largely increases. This leads to high labor costs and the serious problem of incorrect labels (\ie noisy labels). For example, in MSCeleb~\cite{guo2016ms}, the portion of noisy label is beyond 50\% (Fig.~\ref{fig: MSC-R-intra} (a)) which could damage the training. Although some existing work~\cite{deepglint} has refined the dataset (Fig.~\ref{fig: MSC-R-intra} (b)), heavy workload by human is indispensable in general.

To alleviate the workload of human labelling, one of the appropriate solutions is to start with a limited number of labelled data and exploit a large amount of unlabelled data with semi-supervised learning. As deep CNN is data-driven, it is an effective way to boost the training by transferring unlabelled data to labelled ones. However, there are seldom semi-supervised methods besides Consensus-Driven Propagation~\cite{zhan2018consensus} (CDP) proposed for face recognition since the deep learning prevails. The major challenge consists of the incorrect labels introduced by auto-labelling, since the face datasets are generally in large scale which makes the noisy-label problem much more serious. CDP attempts to improve the labelling accuracy by using the committee-mediator mechanism and achieves compelling accuracy on Megaface. However, the error accumulation should be considered especially when the labelling system encounters increasing unlabelled data.

Therefore, to address the noisy label problem that introduced by naive labelling, we propose to draw the advantage of a noisy label learning routine. Certain methods~\cite{lu2015noise,kong2019recycling,ding2018semi} have been proposed to combine the noisy label learning with semi-supervised learning and cope with the noisy labels jointly. They achieve great improvement on the benchmarks of limited scale such as MNIST~\cite{lecun1998mnist} and CIFAR-10~\cite{krizhevsky2009learning}, but are not yet     validated in the case of large-scale face recognition in which the noisy label problem becomes much more serious.
Some other works study to cope with the noisy labels for face recognition~\cite{wu2018light,li2019learning,wang2019co}, but without discussions on semi-supervised learning.

In this paper, we propose a novel multi-agent method, name \textbf{GroupNet} (GN), to conduct robust training on noisy label data; then, benefiting from the robustness to noisy label, we further develop a semi-supervised solution, named \textbf{Noise-Robust Learning-Labelling} (NRoLL), whose functionality includes robust training on small amount of labelled data and accurate labelling on large amount of unlabelled ones. There are two major advantages of our solution. (1) The robustness of GN not only benefits the training on noisy label data, but also improves the accuracy of labelling. (2) If NRoLL keeps labelling more and more unseen data, the accuracy of labelling will increase due to the robust training on accumulated data by NRoLL itself. The convergence is validated by the training-labelling experiments with real-world settings.

In summary, this paper includes three major contributions.
\begin{itemize}\itemsep=-1pt
    \item   We propose a novel multi-agent learning method, \ie GN, to accomplish robust training on noisy label data. Through experimental comparison with the existing noisy label learning methods, our GN shows its superiority by the leading accuracy on various benchmarks even the noise portion is beyond 50\% in the training data.
    \item   We develop a semi-supervised solution, \ie NRoLL, for deep face recognition with few labelled data and full exploitation of unlabelled ones. Resorting to the advantage of GN, our NRoLL is able to not only train network on noisy data robustly but also to conduct accurate labelling. To the best of our knowledge, this is the first attempt of combining semi-supervised learning and noisy label learning to yield state-of-the-art performance on eight benchmarks in the field of deep face recognition. 
    \item   The convergence of NRoLL is validated through the experiment where the labelling accuracy and recognition accuracy both increase along with NRoLL keeps treating more and more unlabelled samples.
\end{itemize}

\begin{figure*}[t] 
    \centering  
    \begin{center}
    \includegraphics[width=0.95\linewidth]{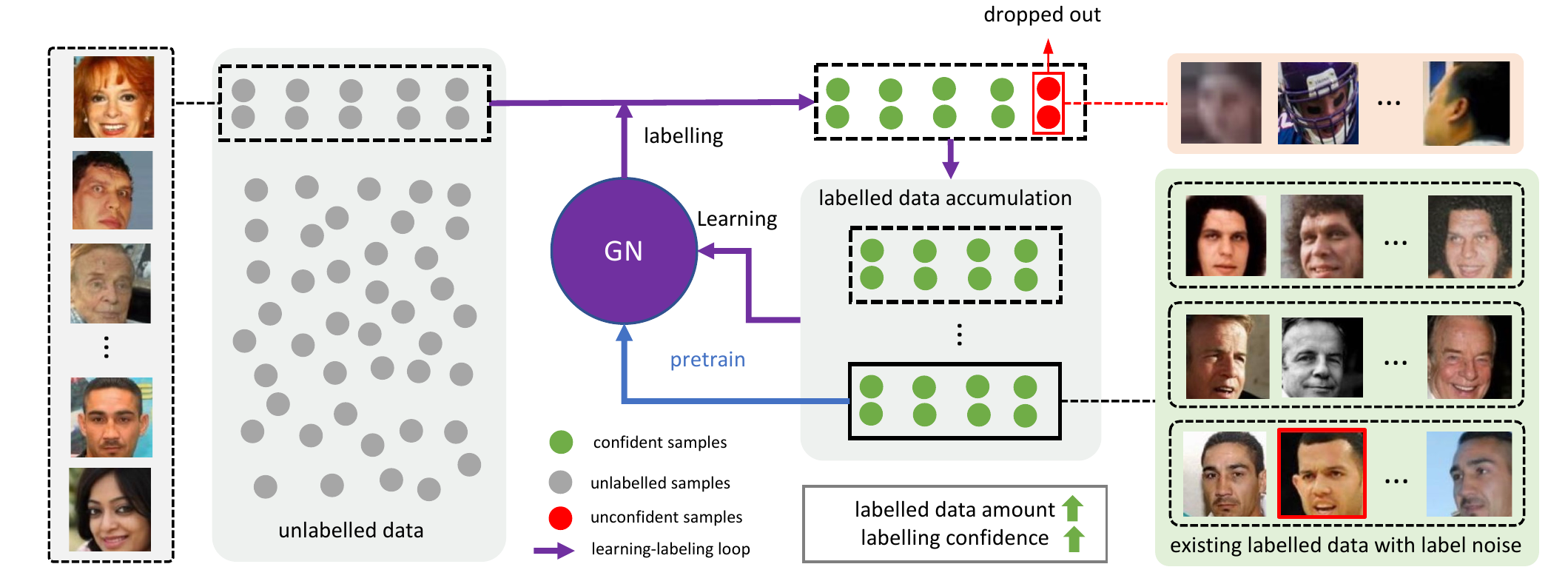} 
    \caption{The overview of NRoLL. 
    GN conducts labelling on the encountered unlabelled data (grey points). Samples with confident predictions are appended to the current labelled dataset, and the unconfident ones (red points) are dropped out. NRoLL repeats the learning and labelling loop when new unlabelled data comes.} 
    \label{fig: NRoLL} 
    \end{center}
\end{figure*}

\section{Related Work}
\label{sec_related}
\subsection{Semi-supervised Learning}
\label{subsec_semi}
There are many research lines of semi-supervised learning techniques, such as generative modelling~\cite{rasmus2015semi,kingma2014semi,lasserre2006principled}.
We focus on the recent popular methods which underpin the semi-supervised image classification. The first widely recognized practice is the consistency regularization. It holds that a model should give consistent predictions on unlabelled samples with small disturbance. This art can smooth models so that they are not sensitive to noise. Approaches in this family includes $\Pi$-Model~\cite{laine2016temporal}, Mean Teacher~\cite{tarvainen2017mean}, Virtual Adversarial Training (VAT)~\cite{miyato2018virtual}, and UDA~\cite{xie2019unsupervised}. Another strategy for semi-supervised learning is to minimize the entropy of predictions for unlabelled data~\cite{grandvalet2005semi}. MixMatch~\cite{berthelot2019mixmatch} combines consistency regularization, entropy minimization and also MixUp~\cite{zhang2017mixup} augmentation to make a holistic solution. 
Self-supervised learning on unlabelled data can also enhance representation learning in semi-supervised learning~\cite{henaff2020data,zhai2019s4l}. Different from auxiliary optimization objective, Pseudo-labelling~\cite{lee2013pseudo} is trained on supervised data. It enlarges labelled data size by assigning confident labels on unlabelled data. This method is suitable for face recognition to deal with massive class numbers and inter-class issues.
However, seldom semi-supervised methods have been validated on the large-scale face recognition task since the deep learning prevails. Recently,  CDP~\cite{zhan2018consensus} is proposed for deep face recognition. It improves the labelling accuracy by using a committee-mediator mechanism. In the experiment, we compare our solution with both CDP and the above representative methods on each face recognition benchmark.

\subsection{Noisy Label Learning}
\label{subsec_noisy}
Research in noisy label learning is flourishing. Certain approaches \cite{goldberger2016training,patrini2017making} estimate the noise transition matrix. But the transition matrix is hard to estimate accurately and efficiently when the class number becomes large. Many recent works focus on the practice of sample selection.  Mentornet~\cite{jiang2018mentornet} pretrains an extra teacher network to select clean samples for the student network. 
Decoupling~\cite{malach2017decoupling} and Co-teaching~\cite{han2018co} train two networks simultaneously. Decoupling selects samples which are predicted differently by the two networks.  Co-teaching considers samples as clean if they have small loss computed by the peer network.  Co-teaching+~\cite{yu2019does} adopts small-loss samples with disagreements between two networks. Co-mining~\cite{wang2019co} identifies clean and noisy faces, and re-weights clean samples and discards noisy ones. 
More recently, a meta-learning based method~\cite{li2019learning} has been proposed for noise-tolerant training.
Besides, certain methods~\cite{lu2015noise,kong2019recycling,ding2018semi} combines noisy label learning with semi-supervised learning to conduct robust training. However, they are only validated on the small-scale tasks such as MNIST and CIFAR-10. Most of them involve two agents to share information. How to make communication between the multiple agents achieve better robust learning is not investigated in face recognition.

\section{Our Method}
\label{sec_method}

In this section, we first introduce GN in Section \ref{sec:groupnet} for robust face recognition with noisy labels. Then, we elaborate the NRoLL solution in subsection~\ref{sec: NRoLL}. 

\subsection{GroupNet}\label{sec:groupnet}
To achieve robust training on noisy label data, GN explores the necessary procedures: 1) discriminate samples into different partitions according to their loss values, and subtly utilize them to indicate the noise level; 2) dispose the communication and information exchange among multiple agents; 3) introduce a novel shuffle strategy to further promote the robustness. 

\begin{figure}[t]
	\centering  
\includegraphics[width=0.8\linewidth]{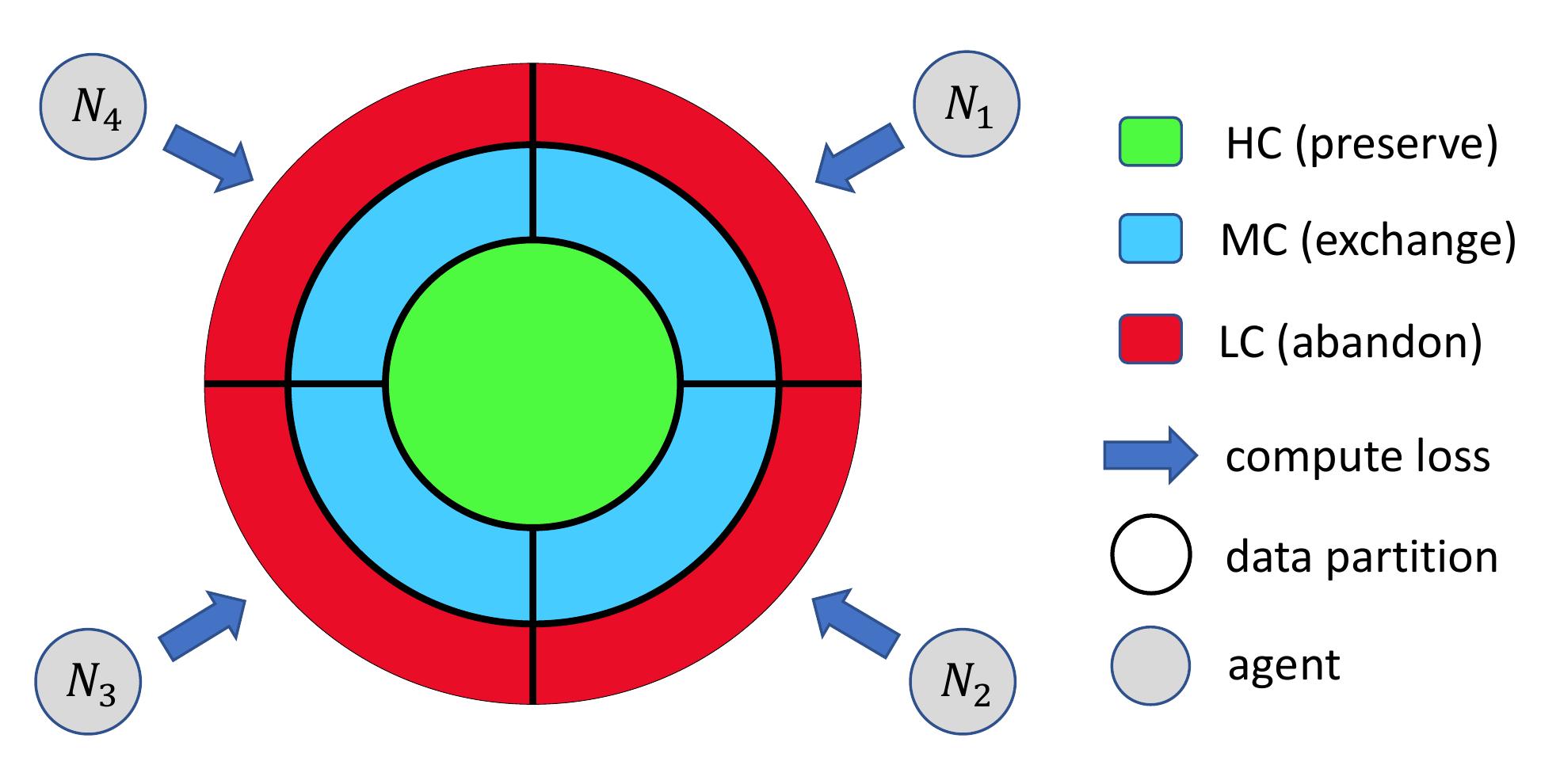}
	\caption{An example of HC, MC, LC (defined in subsection~\ref{sec:groupnet}) discrimination by four agents $N_m, m \in \left \{ 1,2,3,4 \right \}$. The four agents rank a batch of samples in the ascending order of loss values independently. The highest $r\%$ (noise rate) samples are abandoned as $\mbox{LC}$. The intersection between $N_m$ on the remaining samples are identified as HC. The medium partition for each $N_m$ is named $\mbox{MC}$.}
 	\vspace{-1mm}
	\label{fig:HC_MC_LC}
\end{figure}

{\textbf{Discriminate noisy samples.}}
GN employs $M$ peer networks (\ie agents) simultaneously to conduct cooperative discrimination on noisy/clean label data. Specifically, we input the mini-batch to each peer network $N_m, m \in \left \{ 1,2,3, \cdots, M \right \}$, and the forward loss value can be applied to discriminate the samples to three partitions: \textbf{High Confidence (HC)}, \textbf{Medium Confidence (MC)} and \textbf{Low confidence (LC),} as known as noisy sample. In Fig. \ref{fig:HC_MC_LC},
the $M$ agents rank the mini-batch samples in ascending order of loss values independently. The samples with large loss value will be abandoned as LC in red color, whose huge gap with true labels will deteriorate the training. The threshold for LC is determined according to the noise rate $r\%$ estimated from the source training set, \textit{e.g.}, $50\%$ for MSCeleb\cite{guo2016ms}. The intersection between $N_m$ on the remaining samples is identified as HC in green color. 
For example, a sample will be discriminated as HC if and only if it is not discriminated as LC by all the $N_m$.
These HCs usually have very low loss values consistently by multiple agents, which are highly reliable to be utilized for training. 
Excluding LC and HC, the samples in the remaining partition are discriminated by each $N_m$ as MC, which contains clean and potential noisy samples. Note that the MC partitions by $N_m$'s generally have overlap with each other.
For example, a sample is discriminated as LC by $N_1$ and $N_2$, but not by $N_3$ and $N_4$, then it is discriminated as MC by $N_3$ and $N_4$.

\begin{figure}[t] 
    \centering
    \begin{center}
        \subfigure[]{ 
        \begin{minipage}{\linewidth} 
        \centering
        \includegraphics[width = 0.75\linewidth]{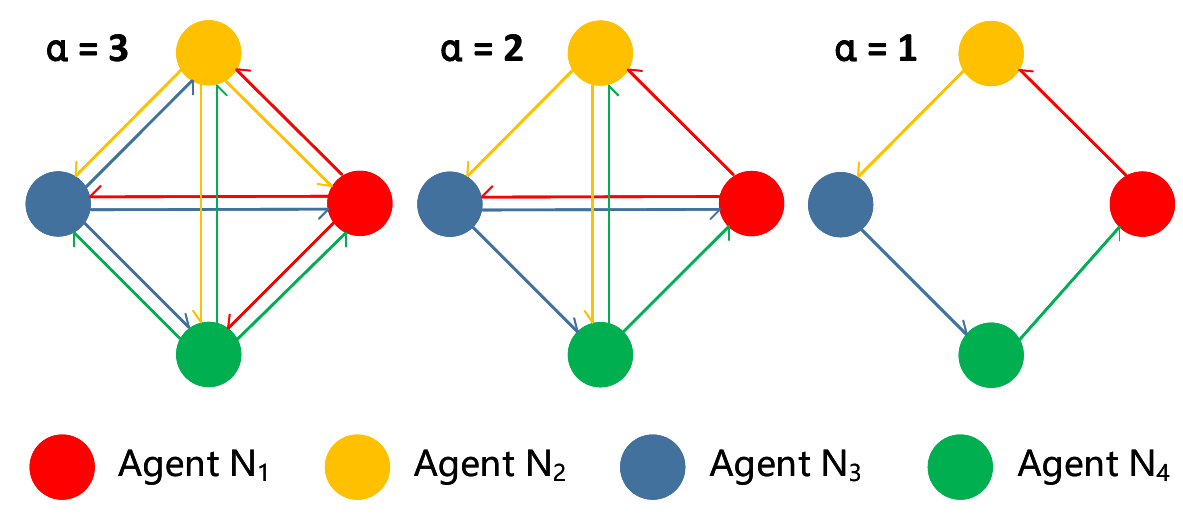} 
        \label{figure: alpha}
        \end{minipage} 
        } 
        \subfigure[]{ 
        \begin{minipage}{\linewidth} 
        \centering 
        \includegraphics[width = 0.75\linewidth]{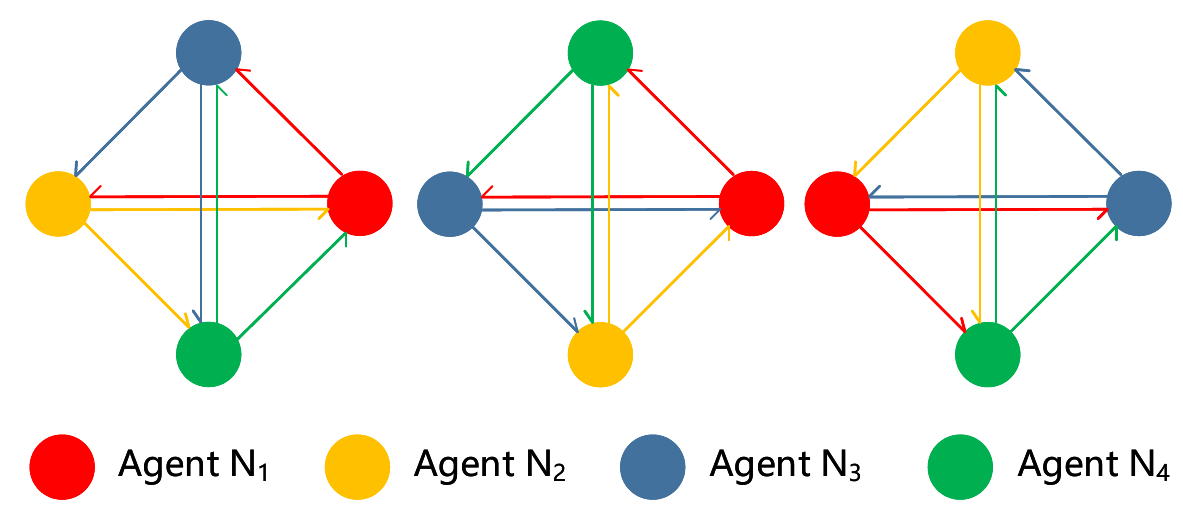} 
        \label{figure: shuffle}
        \end{minipage} 
        } 
    \end{center}
    \vspace{-1em}
    \caption{Information exchange between agents ($M=4$). (a) Each agent broadcasts its MC to the other $\alpha$ agents. (b) Shuffle strategy randomizes the relative positions of agents in a circle in each training iteration.}
    \label{figure: GM}
\end{figure}

{\textbf{Information exchange.}}
To alleviate the error accumulation, the exchange strategy is applied by certain existing works~\cite{han2018co,wang2019co}. 
Compared with the two-agent style, the situation becomes much more complex for multiple agents (\ie the MC exchange among $N_m$). 
Therefore, we develop a novel communication strategy to effectively and efficiently exploit MC samples.
We define a parameter $\alpha \in  \{ 1,2,3, \cdots, M-1  \}$, and each $N_m$ broadcasts its MC to the other $\alpha$ agents along the anti-clockwise direction. As shown in Fig.~\ref{figure: alpha}, we present an example of MC exchange in a GN of $M=4$ agents and different $\alpha$ settings. 
We are able to set $\alpha = M-1$ in maximum, in which each $N_m$ shares MC with all the other peer agents.  
Meanwhile, the recipient should select clean samples from excessive received MC from multiple agents to form a training batch. Here, we adopt the greedy rule to select more confident samples. Specifically, a recipient $N_m$ gives the priority to the samples with higher number of recommendation sources ($\alpha$ in maximum).  For example, in the case of $M=4, \alpha=3$, each agent can recommend and receive MC samples to and from other 3 agents.
Each $N_m$ firstly selects the samples that are recommended (sent) to $N_m$ by $3$ different agents, then selects those recommended by $2$ agents, till the selected samples have the same size with that $N_m$ broadcasts to others. 
For a clear notation, the MC samples selected by the $m$-th agent $N_m$ among all received MC sample from other agents, are notated as $\mbox{MC}_{ms}$.
HC and $\mbox{MC}_{ms}$ samples are both used for training $N_m$ but with different loss functions. Since HC are more reliable than MC, the learning procedure of HC should be emphasised more than that of MC. In implementation, We choose the widely used Arc-softmax~\cite{deng2019arcface} to compute the losses for MC samples while use MV-softmax~\cite{wang2020mis} to compute losses for HC samples.  MV-softmax increases the negative logits in the softmax formula when mis-classification happens and provides stronger supervision than Arc-softmax. For better convergence, the weight for each loss is adaptive according to the size of HC and MC.
The balanced loss of the $m$-th agent, $N_m$, for grouped samples (HC, MC) is:
\begin{align}
\small
\mathcal{L}_{N_m} 
= ~
\frac{1}{\left | \text{HC} \right |+\left | \text{MC}_{ms} \right |}(  
&
\sum_{i=1}^{\left | \text{HC} \right |} \mathcal{L}_{\text{HC}}(N_m(\text{HC}^{i}), y_i) + 
\nonumber \\
&
\sum_{j=1}^{\left | \text{MC}_{ms} \right |} \mathcal{L}_{\text{MC}}(N_m(\mbox{MC}_{ms}^{j}), y_j) 
),
\end{align}
where $\mathcal{L}_{\text{HC}}$ and $\mathcal{L}_{\text{MC}}$ are  MV-softmax loss and Arc-softmax loss, respectively. Note that all agents learn from the same high confidence samples while medium confidence samples ($\mbox{MC}_{ms}$) learned by each agent $N_m$ can be different.


{\textbf{Novel shuffle strategy.}}
Furthermore, we enable each agent to have the opportunity to receive MC from all the other agents during the training procedure even if $\alpha$ is less than $M-1$. 
The purpose is to make the information source diverse and prevent model collapse.
Specifically, we fix the broadcast direction of each location in a circle and shuffle $N_m$ randomly on these locations after each forward propagation. As shown in Fig.~\ref{figure: shuffle}, when $M=4$ and $\alpha=2$, $N_1$ receives MC from one of the three combinations ($N_3$\&$N_4$, $N_2$\&$N_3$, $N_2$\&$N_4$) with the same probability. In this way, each recipient can receive a random combination of MC by this shuffle strategy, when $M$ and $\alpha$ are determined. Note that when $\alpha=M-1$, the shuffle strategy equals to the non-shuffle setting.

{\textbf{Summary.}} According to the loss value of samples, each agent $N_m$ preserves high confidence samples (HC), abandons low confidence samples (LC), and broadcasts medium confidence samples (MC). Each agent trains on the HC and the MC from other agents. Shuffle strategy randomizes the source agents to bring diverse MC recommendations, avoiding error accumulation and model collapse.

\subsection{Noise Robust Learning-Labelling}\label{sec: NRoLL}

In this subsection, we introduce our robust solution for semi-supervised learning in face recognition, named Noise Robust Learning-Label (NRoLL). Taking the advantage of GN, NRoLL starts from training on a small amount of labelled data even the noisy labels exist in it. Then, NRoLL encounters unlabelled data and performs accurate labelling with high confidence. Subsequently, the labelled data grows for further training. The labelling and training boosts each other when NRoLL keeps encountering unlabelled data and converging to increased face recognition accuracy.

{\textbf{Noise Robust pretraining.}} 
Given a small amount of labelled dataset $\mathcal{D}_l$ (not necessarily clean labels), NRoLL firstly makes GN pretrianing on $\mathcal{D}_l$ robustly. Notably, the initial labelled data has two properties. First, the data scale is much smaller than that of the unlabelled data. Second, it is generally with large portion of label noise (\ie 50\% labels are corrupted in MSCeleb). Here, we do not need any human workload for label cleaning because GN is able to conduct robust learning with good tolerance of a large portion of noise. This is the very motivation to use the noise-robust learning method to boost the semi-supervised face recognition.

{\textbf{High confidence labelling.}}
There are numerous unlabelled samples $\mathcal{D}_u$ in real-world (grey dots in Fig.~\ref{fig: NRoLL}). 
These unlabelled faces need heavy labor for the annotation (\ie billions of unlabelled face pictures on the Internet). To simulate the scenario of data accumulation in practice, we set $\mathcal{D}_u$ consists of $S$ parts of unlabelled data $\mathcal{D}_u^t$, where $\mathcal{D}_u = \bigcup_{t=1}^{S} \mathcal{D}_u^t$. NRoLL encounters one part of the unlabelled samples at each time. To obtain reliable labels, each sample in $\mathcal{D}_u^t$ will be predicted by each of the $M$ networks in GN. NRoLL chooses the prediction with the highest logit value as the final prediction given by GN, and the corresponding class is the obtained label. Moreover, NRoLL filters out the unconfident samples if none of the logit values among $M$ predictions overpass the predefined threshold $T$. As shown in Fig.~\ref{fig: NRoLL}, the unconfident samples (red dots) are dropped out during labelling procedure. Finally, NRoLL transfers the current part of unlabelled data $\mathcal{D}_u^t$ to $\mathcal{D}_{pl}^t$ with reliable pseudo labels. 

\begin{algorithm}[t]
\caption{Loop of Noise-Robust Learning \& Labelling }\label{euclid}
\label{alg:NRoLL}
\hspace*{\algorithmicindent} \textbf{Input}: parameters $\left \{ \Theta_1, \Theta_2, \cdots, \Theta_M \right \}$ of $M$ agents in GN,  labelled data $\mathcal{D}_l$ with label noise, unlabelled data  $\mathcal{D}_u =
\bigcup_{i=1}^{S}\mathcal{D}_u^i$, noise rate $r\%$, threshold $T$. \\
\hspace*{\algorithmicindent} \textbf{Output}: parameters $\left \{ \Theta_1, \Theta_2, \cdots, \Theta_M \right \}$, $\mathcal{D}_l$.
\begin{algorithmic}
    \State $i \gets 1$ 
    \State  $r \% \gets \mbox{Expectation-Maximization}(\mbox{GMM}(\mathcal{D}_l))$
    \State $GN\left ( \Theta_1, \Theta_2, \cdots, \Theta_M \right ) \gets$
    \State $~~~~~~~~~~~~~~Train (GN\left ( \Theta_1, \Theta_2, \cdots, \Theta_M \right ), \mathcal{D}_l$, $r\%)$ 
    
    \Repeat 
        
        \State $\mathcal{D}_u \gets \mathcal{D}_u \setminus \mathcal{D}_u^i$
        \State $\mathcal{D}_{pl}^i \gets$ $GN.labelling$($\mathcal{D}_u^i, T$)
        \State ($GN.labelling$ \mbox{uses the class with the highest logit} 
        \State  $~~~~~~~~$\mbox{among all logits ($>T$) given by $M$ agents.})
        \State $\mathcal{D}_l \gets \mathcal{D}_l \cup \mathcal{D}_{pl}^i $
        \State  $r \% \gets \mbox{Expectation-Maximization}(\mbox{GMM}(\mathcal{D}_l))$
        \State  $GN\left ( \Theta_1, \Theta_2, \cdots, \Theta_M \right ) \gets$ 
        \State $~~~~~~~~~~~~Train (GN\left ( \Theta_1, \Theta_2, \cdots, \Theta_M \right ), \mathcal{D}_l$, $r\%$) 
        \State $i \gets i + 1$

    \Until $\mathcal{D}_u = \varnothing $ 
    
    \Return $GN\left ( \Theta_1, \Theta_2, \cdots, \Theta_M \right )$,  $\mathcal{D}_l$, noise rate $r\%$.
\end{algorithmic}
\end{algorithm}

{\textbf{Labelled dataset updating.}} We append $\mathcal{D}_{pl}^t$, \ie the currently encountered part with reliable labels, to the current training data $\mathcal{D}_l$ so that an enlarged training data is obtained.
Importantly, besides the data size, the noise rate of $\mathcal{D}_l$ is also changed from the previous one. The similarity distribution of intra-class face pairs in $\mathcal{D}_l$ is considered as a two-component Gaussian mixture model (GMM) to represent the clean portion and noise portion. Then, we adopt the expectation-maximization (EM) algorithm to estimate the two components. The component with a smaller mean represents the noisy samples and its estimated portion is the noise rate of the updated $\mathcal{D}_l$.

{\textbf{The Loop of NRoLL.}}
After the labelled data $\mathcal{D}_l$ is updated with the incremental set $\mathcal{D}_{pl}^t$, we continue to train GN on the updated $\mathcal{D}_l$ which gains larger data size and less label noise. In the practical face data accumulation scenario, unlabelled data becomes available gradually. When the fresh $\mathcal{D}_u^{t+1}$ is available, NRoLL enables $\mbox{GN}$ to provide reliable labels for $\mathcal{D}_u^{t+1}$ again with the same process introduced above. Consequently, $\mathcal{D}_l$ can be further updated. We formulate the loss function for $N_m$ at the $t$-th NRoLL loop as:
\begin{equation}
\mathcal{L}_{N_m}^{t} = \mathcal{L}_{N_m}(\text{GN}, D_l, D_{pl}^t),
\end{equation}
where
$D_{pl}^t = \bigcup_{i=1}^{\left | D_u^t \right |}\tilde{x}_i$. $\tilde{x}_i = x_i$ if the pseudo label of $x_i$ is  given by GN, otherwise, $\tilde{x}_i = \varnothing$.
This loop keeps working untill no more unlabelled data comes. The whole framework of NRoLL is demonstrated in Fig.~\ref{fig: NRoLL} and Algorithm~\ref{alg:NRoLL}.

{\textbf{Necessity of keeping robustness.}} 
High confidence labelling does not equal to necessary noise proof. Noise in pseudo labels come from misclassified samples and open-set outliers~\cite{wang2018devil}. With the growth of labelled data, the above noise can also be accumulated. This will increase the risk of error accumulation, model collapse and training blow up. Therefore, it is necessary to apply Noise-Robust methods (\eg GN) throughout every learning and labelling loop.
Because of the robustness to noise, the learning and labelling boost each other and lead to better convergence. This will be also discussed in Section~\ref{sec_NRoLL}.

\section{Experiments}

\begin{table}[t]
			\centering
			\caption{Face datasets used for training and testing.}
			\setlength{\tabcolsep}{3mm}{
			\small{
				\begin{tabular}{|c|c|c|c|}
					\hline
					& Datasets &  Images & Identities \\
					\hline
					\hline
					\multirow{2}{*}{Training} & CASIA-Clean~\cite{CASIA-clean} & 0.38M &9,879\\
					& MSCeleb~\cite{guo2016ms}  & 7.03M & 85,173 \\
					\cline{1-4}
					\multirow{7}{*}{Test} & LFW~\cite{LFWTech} & 13,233 & 5,749\\
					& CALFW~\cite{zheng2017cross} & 12,174 & 5,749\\
					& CPLFW~\cite{zheng2018cross} & 11,652 & 5,749 \\
					& AgeDB~\cite{moschoglou2017agedb} & 16,488 & 568 \\
					& CFP~\cite{sengupta2016frontal} & 7,000 & 500 \\
					& RFW~\cite{wang2019racial} & 40,607 & 11,430\\
					& MegaFace~\cite{kemelmacher2016megaface} & 1M & 530 \\
					\hline 
				\end{tabular}
			    }
			}
			\label{tab:datasets}
\end{table}

\subsection{Experimental Settings}
{\textbf{Training datasets.}}
 Two widely used datasets, \ie CASIA-WebFace~\cite{Dong2014Learning} and MSCeleb~\cite{guo2016ms}, are used as training set. We add label noise to a cleaned version of CASIA-WebFace (CASIA-Clean~\cite{CASIA-clean}) to synthesize corrupted data with various noise rate. As for MSCeleb, there are already large amount of noisy labels in it. For experiments conducted on CASIA, we finish the training processes at epoch 20. For experimental efficiency, we finish each training on MSCeleb at epoch 5.

{\textbf{Test datasets.}} 
We evaluate our method on eight popular benchmarks of face recognition: LFW~\cite{LFWTech}, BLUFR~\cite{liao2014benchmark}, CALFW~\cite{zheng2017cross} and AgeDB~\cite{moschoglou2017agedb} with large age gap, CPLFW~\cite{zheng2018cross} and  CFP~\cite{sengupta2016frontal} with large pose variations, RFW~\cite{wang2019racial} with  racial bias and MegaFace~\cite{kemelmacher2016megaface} with million scale of distractors. Details of each training and test set is described in Table~\ref{tab:datasets}.      


\begin{table*}[t]
    \centering
    \caption{Contrast experiments for GN with agent number $M$. The training set is CASIA-Clean with added label noise rate of 30\% and 50\%. GN gives the dominant accuracy on each benchmark even the noise rate reaches 50\% in the training set.}
    \resizebox{\textwidth}{!}{
        \small
        \begin{tabular}{|l|c|c|c|c|c|c|c|c|c|}
            \hline
             Training Set & Agent Num. & LFW & BLUFR$@1e-3$ & BLUFR$@1e-4$ & BLUFR$@1e-5$ & CALFW & CPLFW & AgeDB & CFP \\
            \hline
            \hline
            \multirow{5}{*}   & Baseline & 97.33 & 90.70 & 79.37 & 65.22 & 84  .48 & 74.17	& 85.73	& 81.61 \\
            & $M=2$ & 98.65 & 95.92 & 89.51 & 78.72 & 88.37 & 78.72 & 91.06 & 88.01\\
            CASIA-Clean & $ M=3$ & 98.83 & 96.23 & 90.19 & 79.45 & 88.13 & 78.85 & 91.23 & \textbf{88.37}\\
            Noise = 30\% & $ M=4$ & 98.83 & 96.19 & 90.36 & 79.58 & 88.60 & 79.13 & 90.75 & 87.85\\
            & $ M=5$ & \textbf{98.85} & \textbf{96.37} & \textbf{90.65} & \textbf{79.86} & \textbf{88.92} & \textbf{79.72} & \textbf{91.7} &	88.33 \\
            \hline
            \hline
            \multirow{4}{*} & Baseline & 92.78 & 58.52 & 38.37 & 23.38 & 76.17 & 58.5 & 63.33 & 64.42 \\
            & $ M=2$ & 97.58 & 91.02 & 80.36 & 66.62 & 84.53 & 72.2	& 86.60 & 78.53 \\
            CASIA-Clean & $ M=3$ & \textbf{97.81} & 90.68 & 79.59 & 65.15 & 85.77 & 72.87 &87.57 & 78.67 \\
            Noise = 50\% & $ M=4$ & 97.75 & 91.11 & 80.38 & 66.7	& 85.65	& 72.95 &87.70 & 79.37 \\
            & $ M=5$ & 97.48 & \textbf{91.97} & \textbf{81.91} & \textbf{67.91} & \textbf{85.88} & \textbf{73.62} & \textbf{87.76} & \textbf{80.82} \\
            \hline 
        \end{tabular}
    }
    \label{tab_gn_ablation}
    \vspace{-1mm}
\end{table*}

{\textbf{Data preprocessing.}}
We apply FaceBoxs~\cite{zhang2017faceboxes} to detect faces from raw images and adopt the standard face alignment pipeline~\cite{feng2018wing}. Then, the detected RGB faces are cropped and resized to $120\times 120$. All pixels are normalized to $\left [ -1, 1 \right ]$. Each image has a chance of 0.5 to be horizontally flipped. For a precise evaluation, the overlapped identities between training sets and testing sets are removed according to~\cite{wang2019co}.
        
{\textbf{Baseline and loss function.}} For fair comparisons, all the experiments in this work use MobileFaceNet~\cite{chen2018mobilefacenets} as the CNN architecture. The baseline is a single network trained by Arc-softmax, where the margin $m$ and scale $s$ are set to 0.5 and 32; in GN, we set the hyper-parameter $t$ of MV-softmax~\cite{wang2020mis} loss to 1.1 according to its suggestion.

{\textbf{Training details.}} All methods are trained from scratch with batch size of 512 on four P40 GPUs. We use stochastic gradient descent (SGD) optimizer. The weight decay and the momentum are 0.0005 and 0.9, respectively. 
For the experiments on CAISA, the learning rate is initialized as 0.1, and divided by 10 at the 6K, 9K, 12K iterations, and the training ends at the 15K iterations. 
To train on MSCeleb, the learning rate decays at 28K, 42K, 56K iterations, and we end the training at 70K iterations.
 
{\textbf{Test details.}} Cosine similarity is employed as the metric when comparing face feature pairs. The features are extracted from the last fully connected layer of MobileFaceNet. We follow the official protocol to test on LFW~\cite{LFWTech}, BLUFR~\cite{liao2014benchmark}, CALFW~\cite{zheng2017cross}, CPLFW~\cite{zheng2018cross}, AgeDB~\cite{moschoglou2017agedb}, CFP~\cite{sengupta2016frontal} and RFW~\cite{wang2019racial}. The Cumulative Match Characteristics (CMC) curves and the Receiver Operating Characteristic (ROC) curves are computed on Megaface~\cite{kemelmacher2016megaface}. We randomly choose one agent for testing without any bias.

\begin{figure}[t]
	\centering  
	\vspace{-1em}
	\includegraphics[width=\linewidth]{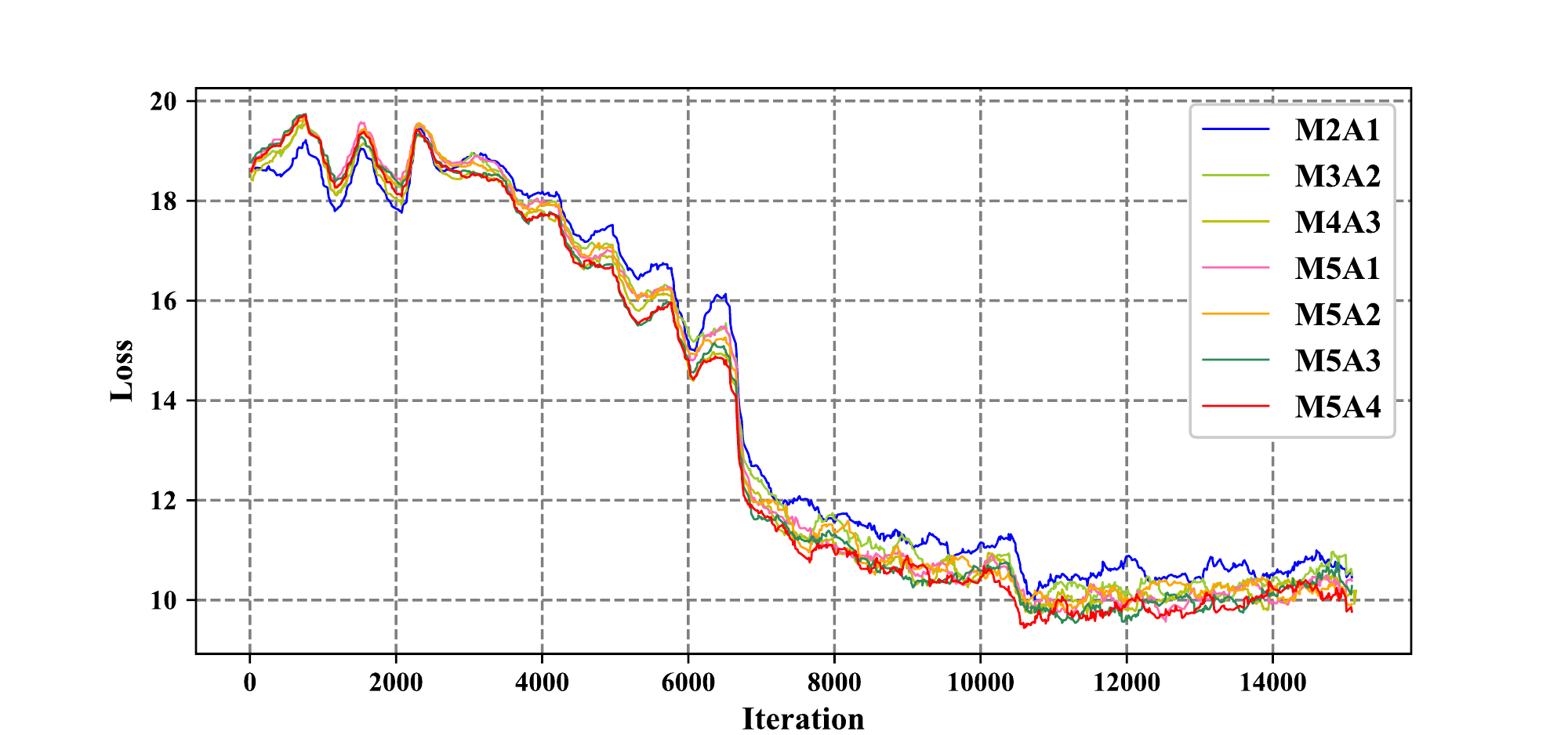}
	\vspace{-1em}
	\caption{Convergence of GN with various setting of $M$ and $\alpha$. Shuffle strategy is applied. 
	They are trained with 50\% label noise added in CASIA-Clean~\cite{CASIA-clean}.
	}
 	\vspace{-1em}
	\label{fig:noisydet}
\end{figure}

\subsection{Effectiveness of GN}     
We evaluate GN on both synthetic noisy data and real-world noisy data. For the concise expression, we use abbreviations like M4A3 to signify GN with $M=4$ and $\alpha=3$ in the following parts and tables.

{\textbf{Convergence and Noise Robustness.}}
Since CASIA-Clean~\cite{CASIA-clean} is a carefully cleaned dataset with hardly noisy labels, it is feasible to synthesize noisy data by adding a precise portion of label noise and conduct contrast experiments on it.
The training loss curves of GN in Fig.~\ref{fig:noisydet} indicate the stable convergence under different parameter settings, even the noise rate reaches 50\%.
In Table~\ref{tab_gn_ablation}, we find that GN by all settings can achieve higher accuracy than the baseline, indicating the robustness to serious label noise. 

{\textbf{Benefits from more agents.}}
Table~\ref{tab_gn_ablation} also shows the performance of GN with different agent number $M$. In most cases, we observe accuracy increases along with the agent number increases. This indicates more agents leads to better robustness.


{\textbf{Impact of $\alpha$ and effectiveness of shuffle strategy.}}
Table~\ref{tab: alpha_shuffle} shows that decreasing $\alpha$ has a negative impact to recognition accuracy when the information exchange is fixed (non-shuffle between agents). By contrast, shuffle between agents can effectively alleviate this issue, especially when $\alpha=1, \alpha=2$. This indicates that the shuffle strategy and increasing $\alpha$ can both diversify information sources and alleviate error accumulation. When connecting all peer agents is not practical in the real scenario, we suggest to play this shuffle strategy. 

\begin{table}[t]
\centering
\caption{ Contrast experiments for GN ($M=5$) with different $\alpha$. The models are trained on CASIA-Clean~\cite{CASIA-clean} with an added noise rate of 50\%. The left underlined digits are the results by GN with shuffle strategy, while the right ones are without shuffle strategy.
}
\small{
\setlength{\tabcolsep}{4mm}{
    \small
	\begin{tabular}{|c|c|c|c|}
		\hline
		\multirow{2}{*}{$\alpha$} & \multicolumn{3}{c|}{BLUFR}\\
		\cline{2-4}
		& $@1e-3$ & $@1e-4$ & $@1e-5$ \\
		\hline
		\hline
		4 & \underline{\textbf{91.97}}/91.97 & \underline{81.91} /81.91 & \underline{67.91}/67.91 \\
		3 & \underline{91.92}/91.64 & \underline{\textbf{82.22}} /81.60 & \underline{\textbf{68.71}}/67.78 \\
		2 & \underline{91.31}/89.67 & \underline{81.15}/77.92 & \underline{67.52}/63.22 \\
		1 & \underline{91.56}/89.80 & \underline{81.13}/77.79 & \underline{67.26}/63.16 \\
		\hline 
	\end{tabular}
}}
\label{tab: alpha_shuffle}
\end{table} 

{\textbf{GN shows advantage on real-world noisy data.}} 
We compare GN with 5 state-of-the-art noisy label learning methods (De-coupling~\cite{malach2017decoupling}, Co-teaching~\cite{han2018co}, Co-teaching+~\cite{yu2019does}, MentorNet~\cite{jiang2018mentornet}, Co-mining~\cite{wang2019co}) and two baselines (softmax and Arc-softmax) on 8 benchmarks. 
All methods are trained on MSCeleb with a large portion of real-world noisy labels (about 50\%).
The experimental results are shown in the middle group of  Table~\ref{tab:all_benchmark}. The major observation is that GN consistently shows its superiority across 8 benchmarks. 
These benchmarks include various types of challenges in face recognition, such as cross pose, cross-age, different races \etc GN outperforms both the supervised baselines and the popular noise-robust methods. 

\textbf{Discussion.} GN requires $m$ times of computational resources compared with the single model system.
If GPU memory is sufficient, multiple agents in GN can run in a parallel way to save time. 
To obtain a robust system and higher performance, it is worthwhile to consume more computational resources,
especially when the computational resource is adequate and higher performance is demanding in the industry.

\begin{table*}[t]
    \centering
    \caption{ Performance (\%) of different methods on 8 benchmarks. 
    The top group includes the supervised learning baseline in face recognition. 
    The middle group includes our GN and the representative methods for noise-robust learning.
    The bottom group includes our NRoll compared with the state-of-the-art competitors for semi-supervised face recognition.
    The training data is MSCeleb (noise rate = 50\%). 
    The middle group trains on the full MSCeleb to compare the robustness to the label noise, while only 1/5 data in MSCeleb are lablled in the bottom group. The bold number refers to the best result of each group.}
    \resizebox{\textwidth}{!}{
        \small
        \begin{tabular}{|l|c|c|c|c|ccc|c|c|c|c|cccc|cc|}
            \hline
            & \multirow{2}{*}{Method} & \multirow{2}{*}{$\#$} &  Training & \multirow{2}{*}{LFW} & \multicolumn{3}{c|}{BLUFR} &  \multirow{2}{*}{CALFW} & \multirow{2}{*}{CPLFW} & \multirow{2}{*}{AgeDB} & \multirow{2}{*}{CFP} &\multicolumn{4}{c|}{RFW} & \multicolumn{2}{c|}{MegaFace}\\
            \cline{6-8} \cline{13-18}
            & & & data & & 1e-3 & 1e-4 & 1e-5 & & & & & C. & I. & As. & Afr. & Rank1 & TPR \\ 
            \hline
            \hline
             
            \multirow{4}{*}{\rotatebox{90}{Supervised}} & Softmax & 1 & 1/5 & 94.63	& 77.14	& 61.53	& 45.85 & 79.30 &   66.22 & 80.10 & 72.00 & 73.43 & 70.25 & 70.30 & 63.97 & 37.48 & 37.64  \\
            & Softmax & 2 &  Full & 98.53 & 95.43 & 87.77 & 73.85 & 87.40 & 76.78 & 88.55 & 89.23 & 85.90 &	78.22 &	77.67 &	74.27 &	67.43 &	69.97 \\
            & Arc-softmax~\cite{deng2019arcface} & 3 & 1/5 & 98.65	& 97.38	& 94.07	& 87.09 & 89.82 & 75.50 & 91.23 & 83.36 & 85.83 & 81.10 & 78.12 & 76.78 & 76.10 & 78.73 \\
            & Arc-softmax~\cite{deng2019arcface} & 4 &  Full & \textbf{99.2} & \textbf{99.13} & \textbf{97.33} & \textbf{91.86} & \textbf{91.62} & \textbf{80.65} & \textbf{93.35}	& \textbf{90.03} & \textbf{91.08}	& \textbf{85.48}	& \textbf{83.68} & \textbf{82.15} & \textbf{83.68} & \textbf{87.19}\\
            
            \hline
            \hline
            \multirow{7}{*}{\rotatebox{90}{Noise-Robust}} & De-coupling~\cite{malach2017decoupling} & 5 & Full & 99.18 & 99.03	& 97.04	& 91.42 & 91.87	& 81.03	& 93.72	& 90.37 & 91.05 & 85.58 & 83.13 & 82.22 & 82.46 & 85.75 \\
            & Co-teaching~\cite{han2018co} & 6 & Full & 99.25 & 99.14 & 97.78 & 93.65 & 91.33 & 76.75 & 93.83 & 83.30 & 91.23 & 85.92 & 83.58 & 82.55 & 82.09 & 85.08 \\
            & Co-teaching+~\cite{yu2019does} & 7 & Full & 99.13 & 98.81 & 96.78 & 91.14 &91.10 & 76.40 & 93.22 & 82.47 & 90.13 & 84.95 & 81.85 & 79.68 & 78.77 & 83.17 \\
            & MentorNet~\cite{jiang2018mentornet} & 8 & Full & 99.25 & 99.10 & 97.37 & 92.02 & 92.03 & 81.55 & 93.35 & 90.41 & 91.38 & 86.25 & 83.88 & 82.83 & 84.30 & 87.41 \\
            & Co-mining~\cite{wang2019co} & 9 & Full & 99.18 & 99.35 & 98.18 & 94.35 & 92.25 & 79.28 & 94.83 & 86.10 & 91.89 & 86.63 & 84.55 & 83.67 & 83.50 & 86.34 \\
            & GN (M4A3) & 10 & Full & 99.35 & 99.37	& \textbf{98.26}	& \textbf{95.39} & 92.35	& \textbf{82.8}	& 94.93 & \textbf{91.07} & 92.83 & \textbf{88.15} & \textbf{85.63} & \textbf{85.15} & \textbf{84.42} & \textbf{87.65} \\
            & GN (M5A4) & 11 & Full & \textbf{99.42}	& \textbf{99.39} & 98.25 & 95.16	& \textbf{92.47} & 82.38 &  \textbf{95.07}	& 90.73 & \textbf{93.18} & 87.75 & 85.58 & 85.00 & 84.40 & 87.23 \\

            \hline
            \hline
            \multirow{10}{*}{\rotatebox{90}{Semi-supervised}}& VAT~\cite{miyato2018virtual} & 12 & 1/5 & 91.68	& 56.62	& 36.99	& 21.71 & 72.55	& 61.23	& 74.32	& 69.11	& 69.67	& 66.77	& 65.93	& 58.03	& 19.89 & 19.46 \\
            & VAT(Arc) & 13 & 1/5 & 97.45 & 93.07 & 84.70 & 71.13 & 86.6 & 70.95 & 86.30 & 79.87	& 80.33	& 76.63	& 72.78	& 68.67 & 61.50 & 62.82 \\
            & UDA~\cite{xie2019unsupervised} & 14 & 1/5 & 87.77 & 38.79 & 24.32 & 14.7 & 67.83 & 57.15 & 68.42 & 61.93 & 64.80 & 63.58 & 62.00 & 57.23 & 11.17 & 11.37 \\
            & UDA(Arc) & 15 & 1/5 & 79.80 & 15.84 & 8.22 & 5.24 & 61.33 & 52.20 & 61.43 & 57.43 & 61.65 & 60.03 & 59.43 & 54.83 & 0.95 & 2.78 \\
            & MixMatch~\cite{berthelot2019mixmatch} & 16 & 1/5 & 76.83 & 11.33	& 6.49 & 4.73 & 61.53	& 52.57 & 59.18 & 55.59 & 61.05 & 58.7 & 59.02 & 54.22 & 2.61 & 0.78 \\
             & Pseudo-Label~\cite{lee2013pseudo} & 17 & 1/5 &97.92 & 94.34 & 86.18 & 71.05 & 86.37 & 73.85 & 86.60 & 85.57 & 82.85 & 77.22 & 76.97 & 71.27 & 61.63 & 65.29 \\
            & CDP~\cite{zhan2018consensus} & 18 & 1/5 & 99.20 & 99.30 & 97.87 & 93.81 & 91.47 & 80.80 & 93.83 & 87.96 & 91.68 & 85.72 & 83.85 & 82.40 & 84.02 & 87.19 \\
            & NRoLL (M3A2) & 19 & 1/5 & \textbf{99.45} & \textbf{99.47} & \textbf{98.6} & 95.74 & 92.28 & 81.88 & 94.83 & 88.09 & 92.65 & 88.4 & 85.67 & 85.27 & 85.45 & 87.96 \\
            & NRoLL (M4A3) & 20 & 1/5 & 99.35 & 99.44 & 98.56 & \textbf{96.19} & \textbf{92.48} & \textbf{81.97} & \textbf{94.85} & \textbf{88.34} & \textbf{92.88} & \textbf{88.48} & 
            \textbf{85.53} & \textbf{85.46} & \textbf{85.77} & \textbf{88.14} \\
            \hline 
        \end{tabular}
    }
    \label{tab:all_benchmark}
    \vspace{-2mm}
\end{table*}

\subsection{Semi-Supervised Experiments of NRoLL}
\label{sec_NRoLL}
To simulate the practical condition of semi-supervised face recognition, we build labelled and unlabelled datasets by splitting MSCeleb. 
Notably, the MSCeleb dataset employed here is always the original version with 50\% of label noise. This brings great challenges to our NRoLL, which encounters two problems (few labelled data and noisy labelled data) in the same time.
Specifically, faces in the same identity are equally split into $S+1$ parts. 
One part is given as the labelled dataset, the rest $S$ parts are deprived of labels to form the unlabelled datasets.
NRoLL will progressively perform noise-robust learning and labelling as described in subsection~\ref{sec: NRoLL}. 
Both the labelled and unlabelled parts come from MSCeleb and are with great portion of noise. Our NRoLL will show its robustness not only for learning but also for labelling.

\begin{figure}[t] 
    \centering
    \begin{center}
        \includegraphics[width=0.9\linewidth]{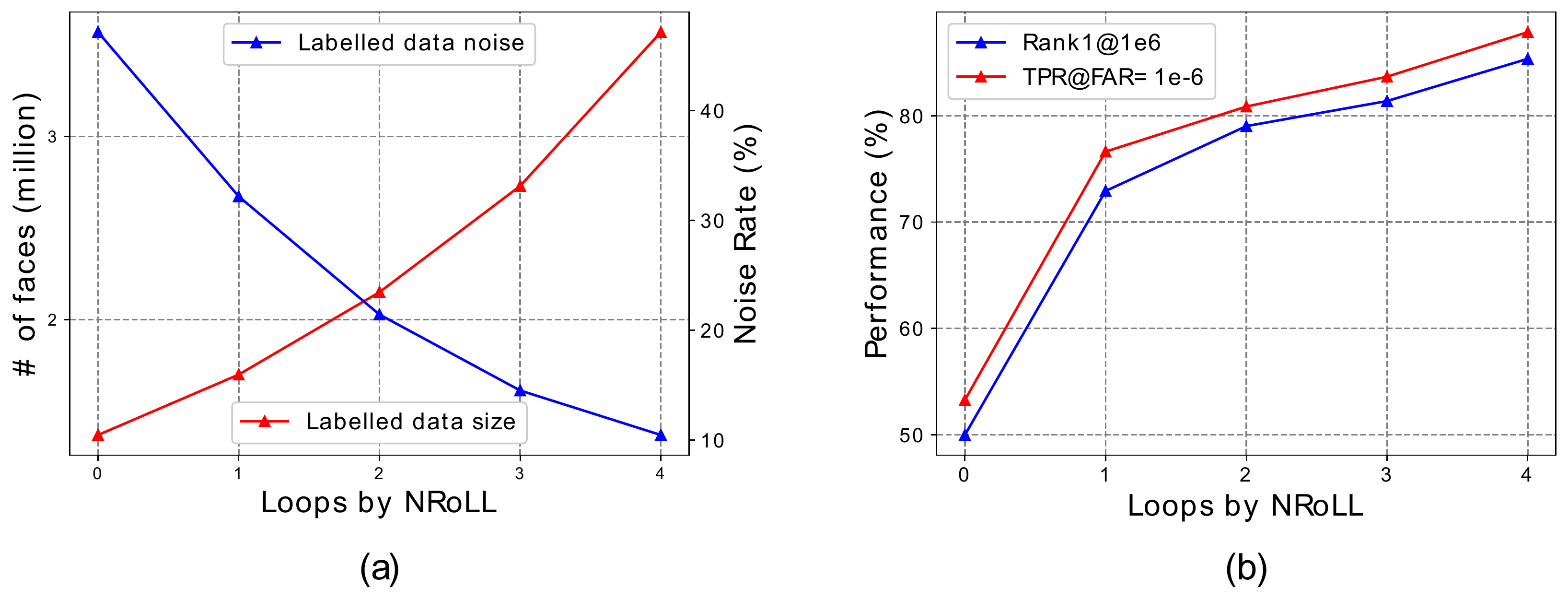}
    \end{center}
    \vspace{-1em}
    \caption{(a) Along with the loop progress by NRoLL, the labelled data size increases, and the noise rate decreases.
    (b) The progressive improvement of testing performance on Magaface along the loop.}
    \label{fig: splt5_status} 
    \vspace{-1em}
\end{figure}

{\textbf{Convergence of NRoLL.}}
We set $S+1=5$ to observe NRoLL running $S = 4$ loops of learning and labelling.
Fig.~\ref{fig: splt5_status} records the state of labelled data (updated in each NRoLL loop) and recognition performance on MegaFace in each loop.
Obviously, the labelled data size grows steadily (red line in Fig.~\ref{fig: splt5_status} (a)).
There are 1.37 million faces in the initial data, while NRoLL can access 3.57 million labelled faces for training after the fourth loop.
Meanwhile, the noise rate of labelled data decreases significantly (blue line in Fig.~\ref{fig: splt5_status} (a)). 
This indicates that we can get a larger and cleaner labelled dataset after each NRoLL loop. 
The quality of labelled dataset is improved gradually, which can bring about training better models.
The performance keeps improving on MegaFace benchmark after each loop. Specifically, rank1 accuracy boosts from 49.97\% to 85.45\%, and precision score finally achieves 87.96\% with 34\% improvements (Fig.~\ref{fig: splt5_status} (b)).  
In summary, the labelled data quality and model performance are improved simultaneously and NRoLL runs towards convergence.
From Fig.~\ref{fig: splt5_status} (b), we can expect further improvement with more data since the performance keeps increasing after the last loop. 
In this progression, GN plays an important role in both learning and labelling. As discussed above, both learning and labelling suffer from the noisy label. This is the major reason that causes the error accumulation and training blow up to the labelling-based methods. Here, NRoLL makes full use of noise-robust advantage from GN and achieves steady and mutual boosting for learning and labelling.



\begin{table}[t]
	\centering
	\caption{Labelling too much data in the same time can impair the dataset size and noise rate of the accumulative labelled data.
	}
	\setlength{\tabcolsep}{3mm}{
	\small{
		\begin{tabular}{|c|c|c|}
			\hline
			Labelling Amount & \multicolumn{2}{c|}{Details of Labelled Data}\\
			\cline{2-3}
			in Each Loop & Dataset size (million) & Noise rate (\%) \\
			\hline
			\hline
			1 part  & 3.57 & 10.46 \\
			2 parts  & 3.38 & 11.31 \\
			4 parts  & 2.69 & 13.53 \\
			\hline 
		\end{tabular}
	}}
	\label{tab:labelling_amount}
			\vspace{-1mm}
\end{table}

\begin{figure}[t]
    \centering
    \begin{center}
        \vspace{-0.5em}
        \includegraphics[width=3.0in]{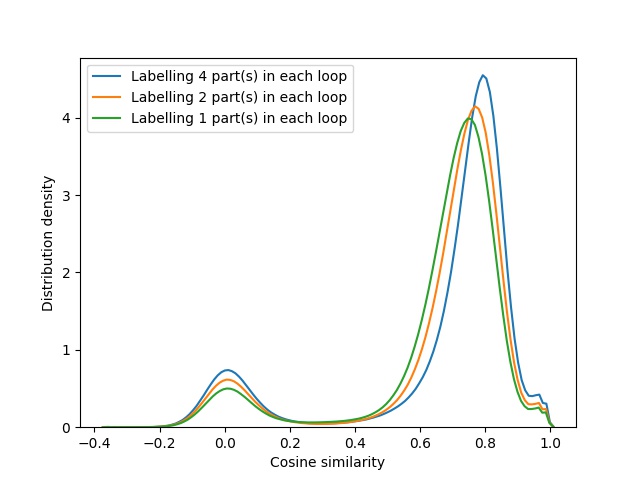}
        \vspace{-1em}
        \caption{Normalized distribution of intra-class face pair similarities after the last loop of NRoLL. $S$ is set to 5 (1 part of labelled data and 4 parts of unlabelled data). NRoLL chooses to label 1 part, 2 parts, or 4 parts of unlabelled data in each loop. 
        } 
        \label{fig: labelling_schedule} 
    \end{center}
    \vspace{-1em}
\end{figure}

{\textbf{Influence of the labelling amount in each loop.}} 
Similarly, MSCeleb is equally split into 5 parts (1 labelled and 4 unlabelled).
We adopt 3 types of labelling amount (1 part, 2 parts, 4 parts) per loop.
Although the total data amount is unchanged, the noise rate and dataset size of the final labelled data gets different.
According to Table~\ref{tab:labelling_amount}, labelling the 4 parts in the same time leads to less labelled data (2.69 million) and higher noise rate ($13.53\%$).
We also use Fig.~\ref{fig: labelling_schedule} to displays the distribution of intra-class face pair similarities after NRoLL deals with all unlabelled data. Intra-class pair similarities distributions are approximated by two-component GMM. 
Two main trends we can get from Fig.~\ref{fig: labelling_schedule}.
1) The ratio between the under curve area of the left peak (noise) and the whole under curve area decreases when NRoLL chooses a smaller labelling amount.  
2) When labelling unlabelled data at once (blue curve), NRoLL can only label some easy samples with high intra-class similarities (around 0.8). On the other hand, if we reduce labelling amount in each loop into 1 part (green curve), NRoLL can assign accurate labels to diverse samples with smaller intra-class similarities (around 0.7).   

\begin{figure*}[t]
    \centering
    \vspace{-1em}
    \begin{center}
        \subfigure[$S=3$]{
        \includegraphics[width=1.65in]{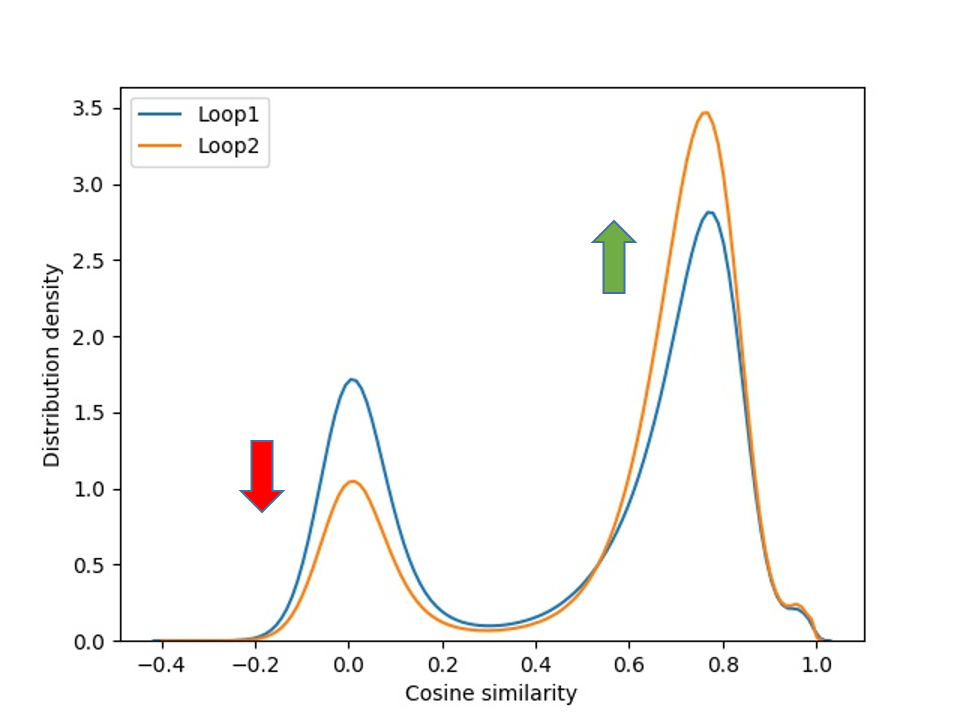}}
        \subfigure[$S=4$]{
        \includegraphics[width=1.65in]{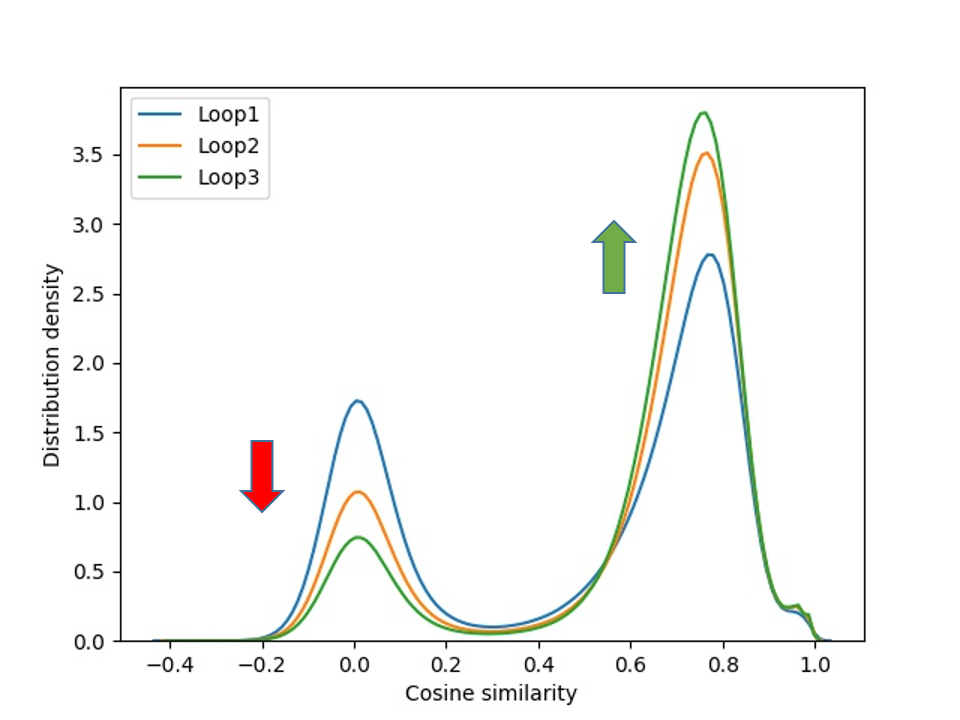}}
        \subfigure[$S=6$]{
        \includegraphics[width=1.65in]{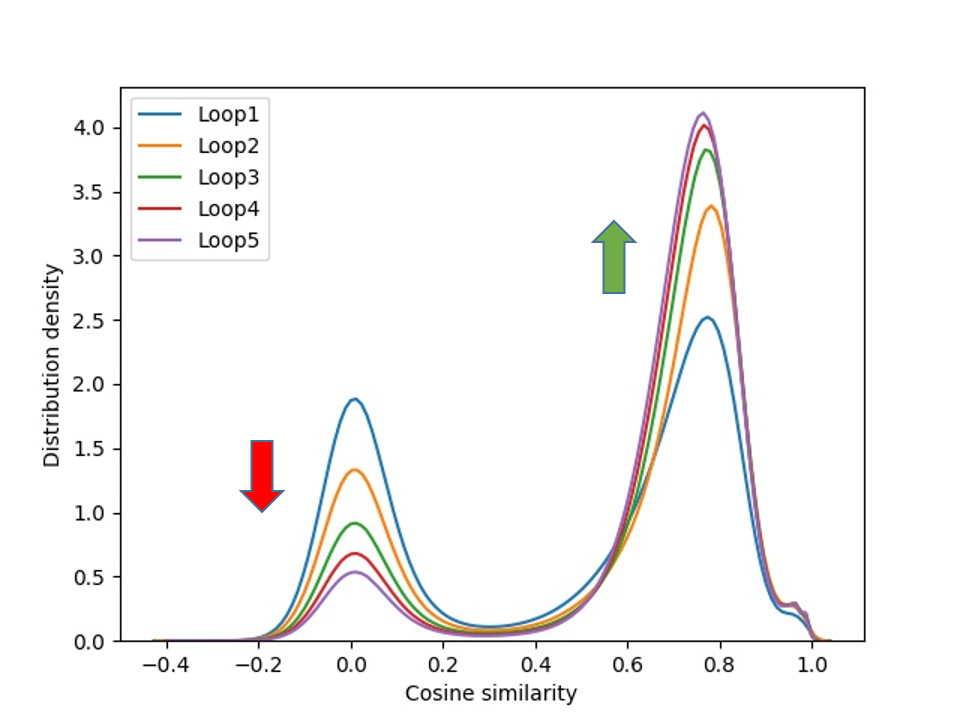}}
        \subfigure[$S=9$]{
        \includegraphics[width=1.65in]{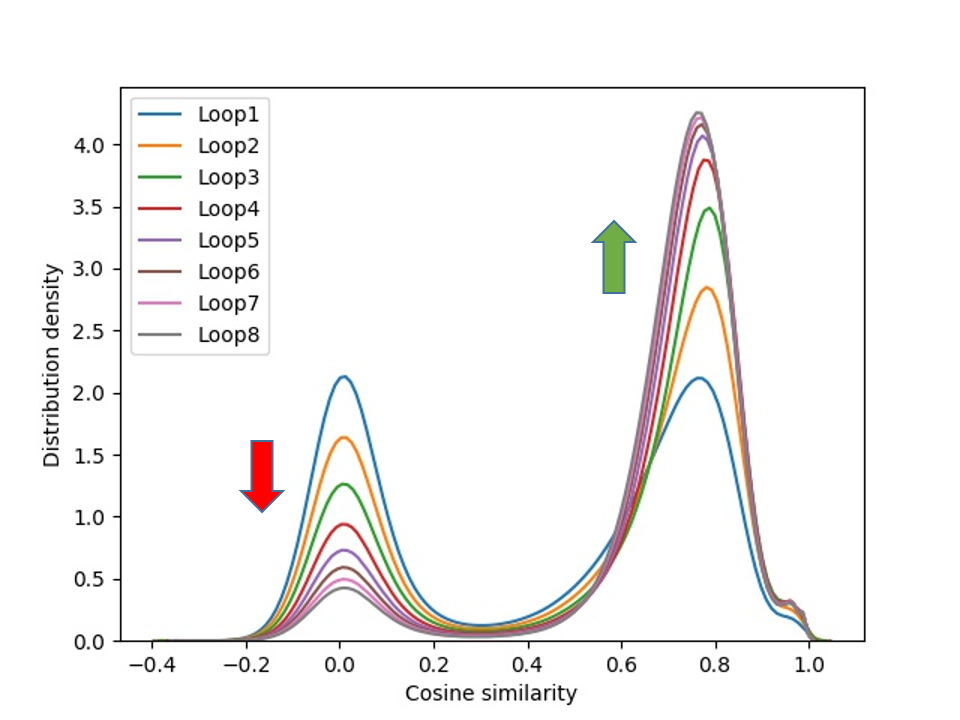}}
        \caption{Normalized distribution of intra-class face pair similarities. Each sub-figure refers to an experiment of NRoLL. NRoLL is pretrained on $1/S$ noisy MSCeleb data and encounters $1/S$ unlabelled data in each loop. The similarities of noisy intra-class pairs are with small cosine values (left peak). Along with the NRoLL procedure, the ratio of noisy samples is decreasing (red arrow) and clean samples are being accumulated into the labelled data (green arrow).} 
        \label{fig: different_slipt} 
    \end{center}
    \vspace{-1em}
\end{figure*}

{\textbf{Comparison with other methods.}}
We implement a set of state-of-the-art methods for semi-supervised face recognition, including VAT~\cite{miyato2018virtual}, UDA~\cite{xie2019unsupervised}, MixMatch~\cite{berthelot2019mixmatch}, Pseudo-Label~\cite{lee2013pseudo}, CDP~\cite{zhan2018consensus} and compare them with NRoLL on 8 benchmarks in Table~\ref{tab:all_benchmark}. 
VAT, UDA and MixMatch are the state-of-the-art semi-supervised learning methods on the classic machine learning benchmarks (\ie SVHN~\cite{netzer2011reading}, CIFAR-10~\cite{krizhevsky2009learning} and ILSVRC-2012~\cite{russakovsky2015imagenet}). 
However, for the face recognition task with extremely noisy labels and large-scale identities, all of them perform under expectation. 
Even compared with the plain supervised learning baselines (\# 1, 3), the performance gap of the above methods is significant. 
CDP~\cite{zhan2018consensus} is recently proposed with a KNN based method to label positive face pairs based on pre-trained face embedding spaces.
We employ 4 diverse committees and 1 mediator in the re-implemented version. As CDP is not functionally robust to noisy labels during training, the performance of CDP (\#18) on 1/5 labelled MSCeleb can not achieve that of NRoLL (\# 19, 20), even the agent number of NRoLL is reduced to 3 with smaller parameter size. On the other hand, the labels given by CDP can not be merged with the original labelled dataset where CDP committees are trained. However, NRoLL can naturally achieve such data accumulation as shown in Fig.~\ref{fig: NRoLL}.
NRoLL can also beat all noise label learning methods (\# 5-9) and baselines (\# 1-4) on most benchmarks, even they are trained with the full supervision of MSCeleb.
Compared with GN (\#10,11) trained on full MSCeleb, NRoLL with GN is still competitive even only 1/5 labelled data is available. 

\begin{table}[t]
	\centering
	\caption{Performance (\%) of NRoll on Megaface with different labelled data ratios of MSCeleb.}
	\small{
	\setlength{\tabcolsep}{1.5mm}{
		\begin{tabular}{|c|c|c|c|c|c|c|}
			\hline
			\multirow{2}{*}{MegaFace Test} & \multicolumn{6}{c|}{Labelled MSCeleb}\\
			\cline{2-7}
			& 1/3 & 1/4 & 1/5 & 1/6 & 1/7 & 1/9 \\
			\hline
			\hline
			Rank1@1e6 & 84.59 & 85.04& \textbf{85.45}& 84.61  & 84.95 & 84.87\\
			TPR@FAR=1e-6 & 86.98 & 86.89 & \textbf{87.96} & 87.57 & 86.48 & 86.79\\
			\hline 
		\end{tabular}
	}
	}
	\label{tab:MSCeleb}
	\vspace{-3mm}
\end{table}

{\textbf{Robustness to few labelled data.}} 
Here, we conduct experiments for NRoLL with fewer labelled data, shown in Table~\ref{tab:MSCeleb}.
The noisy MSCeleb is equally divided into $S$ parts ($S = 3, 4, 5, 6, 7, 9$), and only one part (1/$S$ data) is given as labelled. NRoLL encounters one part of unlabelled data in each loop, so $S-1$ loops are repeated. The testing results on MegaFace show that with only $1/9$ labelled data, NRoLL can achieve similar performance with that gained from $S=3$ (\ie Rank1$=84.87\%$, TPR@FAR$=86.79\%$). 
Therefore, we believe NRoLL is able to perform effective and cost-efficient face recognition in practical scenarios with the rough conditions.
We also visualise the distribution of intra-class face pair similarities labelled by NRoLL after each loop in Fig.~\ref{fig: different_slipt}. The area under each curve is normalized to 1. The similarity distribution of intra-class face pairs can be approximated by a two-component Gaussian mixture model (GMM). The left component with the smaller mean is computed from the noisy intra-class pairs, whereas the clean intra-class pairs contribute to the right peak with the larger mean. When increasing NRoLL loops, faces with clean labels are accumulated progressively. This trend is consistent with various $S$.
Even only $1/9$ data is accessible as labelled at the begging (As shown in Fig.~\ref{fig: different_slipt}(d)), NRoLL is still able to update the labelled data to a much cleaner version after 8 loops. 

{\textbf{Robustness to the open-set scenario.}} NRoLL can be easily extended to the open-set semi-supervised face recognition scenario, where the identities in the unlabelled data do not necessarily appear in the labelled set. To mimic the real semi-supervised learning scenario of face recognition, we assume both the existing identities in the labelled data and the new identities should appear in the unlabelled data. We equally split MSCeleb (around 50\% label noise) into 3 parts with the same number of faces. The first part includes labelled samples and covers half of the total identities, while the other two parts are unlabelled. For each loop in NRoLL, we use the feature of the unlablled face as the prototype of a new identity if the cosine-similarity between the unlablled face and all existing identities is smaller than a threshold. The prototype $F$ is updated by the moving average mechanism in the $i$-th step:
\begin{equation}
F_{i} = \alpha \times F_{i-1}  + (1- \alpha) \times F_{i},
\end{equation}
where $\alpha$ is set as 0.9 in our experiments.
\begin{table}[t]
	\centering
	\caption{Performance (\%) on LFW and BLUFR with different labelling-based methods.}
	\small{
	\setlength{\tabcolsep}{3.3mm}{
	    \small
		\begin{tabular}{|c|c|c|c|c|}
			\hline
			\multirow{2}{*}{Methods} & \multirow{2}{*}{LFW} & \multicolumn{3}{c|}{BLUFR}\\
			\cline{3-5}
			& & 1e-3 & 1e-4 & 1e-5 \\
			\hline
			\hline
			Pseudo-Label~\cite{lee2013pseudo} & 99.20 & 99.19 & 97.54 &92.87 \\
			NRoLL & 99.21 & 99.33	& 98.43	& 94.88\\
			\hline 
		\end{tabular}
	}
	}
	\label{tab:oepset_1}
	\vspace{-1mm}
\end{table}

\begin{table}[t]
	\centering
	\caption{Performance (\%) on RFW and Megaface with different labelling-based methods.}
	\small{
	\setlength{\tabcolsep}{1.3mm}{
	    \small
		\begin{tabular}{|c|c|c|c|c|c|c|}
			\hline
			\multirow{2}{*}{Methods} & \multicolumn{4}{c|}{RFW} & \multicolumn{2}{c|}{MegaFace}\\
			\cline{2-7}
			& C. & I. & As. & Afr. & Rank1 & TPR \\
			\hline
			\hline
			Pseudo-Label~\cite{lee2013pseudo} & 91.15 & 85.23 & 82.58 & 82.13 & 82.74 & 86.36 \\
			NRoLL & 91.62	& 87.4 & 84.93 &84.47 & 85.99 & 88.41 \\
			\hline 
		\end{tabular}
	}
	}
	\label{tab:oepset_2}
\end{table}
We compare NRoLL with the conventional Pseudo-Label~\cite{lee2013pseudo} method on different benchmarks. As shown in Tale~\ref{tab:oepset_1} and Table~\ref{tab:oepset_2}, NRoLL can achieve promising improvement over the conventional pseudo label method.
We can conclude that GroupNet and NRoLL framework can still work robustly in the open-set semi-supervised face recognition with significant label noise.  

\section{Conclusions}

In this paper, we propose a novel noisy label learning method GN and take its advantage into our semi-supervised solution NRoLL for deep face recognition. Benefiting from the robustness to noisy labels and the exploitation of unlabelled data, one can not only train high-performance networks for face recognition from a small amount of labelled data, but also gain extra labelled data from unlabelled ones with increasing accuracy of labelling. To the best of our knowledge, it is the first attempt of combining noisy label learning and semi-supervised learning for deep face recognition with leading accuracy on extensive benchmarks. 


\ifCLASSOPTIONcaptionsoff
  \newpage
\fi



%

\bibliographystyle{IEEEtran}
\bibliography{reference}

\end{document}